\documentclass[letterpaper]{article} 
\usepackage{aaai2027}  
\usepackage{amsmath,amssymb,amsthm,mathtools,bm}
\newcommand{\errfont}{\fontsize{4.5}{5}\selectfont}
\newcommand{\stat}[2]{\(#1\){\errfont\,\(\pm #2\)}}
\newcommand{\beststat}[2]{\(\mathbf{#1}\){\errfont\,\(\mathbf{\pm #2}\)}}
\newcommand{\awcell}[2]{\begin{tabular}[c]{@{}c@{}}#1\\[-0.45ex]#2\end{tabular}}
\providecommand{\missingcell}{\textemdash}

\usepackage[hyphens]{url}  
\usepackage{graphicx} 
\usepackage{natbib}  
\usepackage{caption} 
\usepackage{multirow}
\usepackage{booktabs}
\usepackage{makecell}
\usepackage{cuted}
\usepackage{caption}
\usepackage{algorithm}
\usepackage{algorithmic}

\newtheorem{theorem}{Theorem}

\newcommand{\method}{PopFS}

\usepackage{newfloat}
\usepackage{listings}
\DeclareCaptionStyle{ruled}{labelfont=normalfont,labelsep=colon,strut=off} 
\floatstyle{ruled}
\newfloat{listing}{tb}{lst}{}
\floatname{listing}{Listing}

\usepackage{booktabs}

\title{Population-Robust Feature Selection via Generalized Welfare Optimization}
\author{
    Ruiqi Lyu\textsuperscript{\rm 1}\corresponding,
    Alistair Turcan\textsuperscript{\rm 1},
    Bryan Wilder\textsuperscript{\rm 1}
}

\affiliations{
    \textsuperscript{\rm 1}Carnegie Mellon University\\
    * ruiqil@andrew.cmu.edu
}

\begin{document}

\maketitle
\begin{abstract}
Choosing which features to collect is a deployment decision: the same limited questionnaire, test panel, or sensor set may need to serve several heterogeneous populations. Standard feature-selection methods typically optimize for one large population, while existing robust approaches tend to learn one shared model for every population. We introduce \method{}, a method for learning one shared, deployable feature set that is robust to population differences while letting each population train its own model. \method{} uses a tunable welfare objective that lets practitioners balance overall predictive benefit against stronger protection of the populations that benefit least. To make this objective practical at scale, \method{} first uses multitask sparse learning to reduce the candidate pool, then searches directly over hard feature sets by ranking promising additions and swaps and fully refitting only a shortlist. Across eight population splits from six prediction tasks drawn from five tabular and public-health datasets, \method{} consistently achieves strong average and worst-population performance while scaling to thousands of candidate features. A 43-state COVID-19 nowcasting study further shows that changing the welfare objective can improve the least-served states with little change in average performance and yields an interpretable change in the selected symptom signals. Our code is available at \url{https://github.com/Rachel-Lyu/PopFS}.
\end{abstract}

\section{Introduction}

Many machine-learning systems are developed using richer data than can be collected once the system is deployed \citep{saar2006active,kapoor2005learning}. A pilot health study may record hundreds of survey responses and clinical measurements, while a deployed screener must use only a short questionnaire or test panel \citep{kroenke2003patient}. Similar constraints arise in sensor placement, credit assessment, and other settings where each additional measurement carries a financial, operational, privacy, or human cost \citep{guestrin2005near}. The challenge becomes sharper when the same measurements must be collected across several populations that differ in distribution \citep{sagawa2019distributionally,hashimoto2018fairness}. A health system, for example, may require every clinic to administer the same screening protocol across heterogeneous patient populations, for some of which fitting a reliable separate model is itself challenging. Specifically, from $p$ candidate features, we must choose exactly $k$ features, where $k \ll p$, that support useful prediction for each population. After the shared feature set is fixed, each population may deploy its own student model, chosen to match local sample size, modeling constraints, and operational requirements. A feature set selected only for pooled accuracy may work well on average while preserving little useful information for populations whose predictive relationships differ from those of the majority.

Existing work addresses important parts of this problem, but less so their combination. Classical feature selection generally asks which variables best support prediction for one distribution or one fitted model \citep{guyon2003introduction,kohavi1997wrappers,tibshirani1996regression}. Distributionally robust and group-robust learning instead protect performance across distributions, but typically do so by fitting one shared predictor \citep{duchi2021learning,sagawa2019distributionally,
hashimoto2018fairness}. The closest work to our setting is Distributionally Robust Feature Selection (DRFS), which selects one shared feature set via a continuous relaxation to protect the worst-performing population while allowing a different downstream model in each population \citep{swaroop2026distributionally}. This leaves two practical gaps. First, robustness is not one-size-fits-all. Some applications prioritize overall benefit, while others are willing to trade some average performance for stronger protection of the populations that benefit least. Second, feature acquisition is ultimately a discrete design decision: a system deploys a particular questionnaire, test panel, or sensor set, not a collection of partially included features. Optimizing a continuous relaxation and rounding it afterward can therefore create a gap between the objective being optimized and the feature set that is ultimately deployed. We seek a method that makes the population trade-off explicit and tunable, while enabling a scalable search over very large feature sets.

We introduce \method{}, a scalable, tunable method for learning a robust shared feature set while fitting a separate prediction model in each population. \method{} makes the trade-off between overall benefit and protection of lower-benefit populations explicit through a simple welfare hyperparameter. \method{} achieves high accuracy through directly searching over discrete feature sets instead of continuous objectives, while making this tractable through a series of pre-screening and surrogate measures which can approximate the discrete objective. Across multiple population splits in diverse settings, \method{} improves in both average and worst-case population performance by up to \(22\%\) over baselines while running in $<15$ minutes regardless of total feature size. In a COVID-19 nowcasting case study, tuning the welfare hyperparameter of \method{} enables up to \(40\%\) higher performance for the least-served states without sacrificing performance overall. Our contributions are as follows:
\begin{enumerate}
    \item We introduce \method{}, a method for learning one shared deployable feature set that is robust to population differences under a tunable welfare objective.
    \item We develop a scalable optimization strategy that directly searches over discrete feature sets using multitask sparse learning for screening and a Gauss--Newton-ranked refit search.
    \item \method{} achieves strong performance in population-average and worst-population comparisons and demonstrates interpretable shared feature selection in a COVID-19 application.
\end{enumerate}

\section{Related work}

\paragraph{Feature selection.}
Feature-selection methods include filters and wrappers that search candidate subsets and embedded methods that induce sparsity during model fitting \citep{guyon2003introduction,kohavi1997wrappers,tibshirani1996regression}. Exact cardinality-constrained selection is discrete even for linear prediction \citep{natarajan1995sparse}, motivating greedy approximations and sparse regularization \citep{das2011submodular,khanna2017scalable,tibshirani1996regression}. These methods generally assume one target distribution and assess features through a particular model's performance or coefficients. Multi-task feature learning allows task-specific predictors to share a common feature set \citep{argyriou2006multi,obozinski2010joint,yuan2006model}, but typically selects that set by minimizing aggregate prediction loss across tasks rather than explicitly balancing the benefit received by each population. Sample- or cluster-specific masks address a different acquisition setting because individuals may receive different measurements \citep{yang2022locally,svirsky2023interpretable}. Full-feature reference models also relate to knowledge distillation and privileged information \citep{hinton2015distilling,lopez2015unifying}. \method{} instead learns a shared feature set for deploying unique models to each population while considering the balance between these populations' performance.

\paragraph{Distributionally robust optimization.}
Distributionally robust optimization seeks predictors stable across distributions, while group DRO protects the worst specified population \citep{duchi2021learning,sagawa2019distributionally,hashimoto2018fairness}. These methods typically fit one shared predictor, so sparsifying it yields a robust sparse model rather than a shared acquisition set for separate population-specific models. DRFS introduces this shared-acquisition setting, selecting one feature set while allowing population-specific downstream predictors \citep{swaroop2026distributionally}. \method{} generalizes DRFS's worst-population objective: it provides a tunable hyperparameter to vary aggregate benefit and protection of lower-benefit populations. DRFS relaxes binary inclusion and learns the least-noised features after continuous optimization. \method{} instead searches directly over deployable hard subsets while staying computationally tractable, which is more directly aligned with the discrete selection objective. Concurrent safe distributionally robust feature selection seeks a superset containing every support that may become optimal under a range of covariate shifts \citep{hanada2026safe}, a coverage objective complementary to selecting one welfare-optimal set for observed populations.


\section{Problem formulation}
\label{sec:problem}
Our aim is to learn one set of shared features for all \(r\) populations such that each population can train an appropriate deployed student model on those features. Population $i\in[r]$ has distribution $P_i$ over $(X,Y)$, where $X=(X_1,\ldots,X_p)$ comprises $p$ candidate features. Here, \([m]=\{1,\ldots,m\}\), population \(i\) contributes training observations \(\{(x_{it},y_{it})\}_{t=1}^{n_i}\), and \(X_S=(X_j)_{j\in S}\) for any \(S\subseteq[p]\). 
The goal is to choose one shared subset $S\subseteq[p]$, $|S| = k$, used by a separate prediction model in every population, that will not sacrifice worst-case performance in favor of the average case, and vice-versa, while providing a tunable balance between these. The difficulty is that a feature set that preserves the predictive signal in one population may discard it in another, but it is computationally intractable to learn appropriate feature sets by fitting models on every possible subset in every population.



\section{Method}
\label{sec:method}
\paragraph{Overview.}
\method{} selects one shared feature set while allowing each population to fit its own predictor. In each population, we first fit a model using all available features; its predictions estimate the predictive signal that the selected features should retain. To evaluate a candidate set, we fit a population-specific model using only those features and assess how well it reproduces the full-feature model's predictions. This teacher objective is the default; optional label and mixed objectives instead use label utility or a weighted combination of teacher and label utilities.
During selection, population-specific improvements over no-feature baselines are combined through a tunable welfare objective, ranging from average benefit to stronger protection of lower-utility populations. After the shared feature set is selected, each population may refit its deployed model to its own outcomes. Directly evaluating every subset would require fitting population-specific models for every combination of features, so \method{} first uses multitask sparse learning to screen features, then uses Gauss--Newton surrogates to shortlist feature sets for full validation refits. Thus, screening and ranking make the search scalable, while validation welfare determines the deployed subset.

\subsection{Teacher-Compression Objective}
\label{sec:teacher_welfare}

To evaluate the performance of one shared feature set, we ask how much of the predictive signal available from all features can still be recovered using the selected features. This separates information lost through feature restriction from irreducible variation in the observed outcomes: under squared loss, only the error in approximating the full-feature conditional mean depends on the subset. Because this conditional mean is unknown, we later approximate it with a fitted full-feature teacher and aggregate the resulting population-specific benefits through a tunable welfare objective. Let \(\omega(Y)\in\mathbb{R}^q\) denote the response representation and assume it is square-integrable under each \(P_i\). Define \(m_i(x)=\mathbb{E}_{P_i}\{\omega(Y)\mid X=x\}\), which is also square-integrable. For \(S\subseteq[p]\), let \(\mathcal X_S\) be the domain of \(X_S\), write \(\mathcal X=\mathcal X_{[p]}\), and let \(\mathcal G_{i,S}\) be a nonempty class of measurable predictors \(g:\mathcal X_S\to\mathbb R^q\) available to population \(i\) when only \(X_S\) is observed. The classes \(\{\mathcal G_{i,S}\}_S\) are induced by the same population-specific modeling rule under different feature restrictions. Assume \(g(X_S)\) is square-integrable for every \(g\in\mathcal G_{i,S}\). Theorem~\ref{thm:projection} formalizes why approximating \(m_i(X)\) is the relevant target for feature selection.

\begin{theorem}[Full-signal decomposition]
\label{thm:projection}
For every population \(i\), subset \(S\subseteq[p]\), and \(g\in\mathcal G_{i,S}\), $\mathbb{E}_{P_i}\|\omega(Y)-g(X_S)\|_2^2 = \mathbb{E}_{P_i}\|\omega(Y)-m_i(X)\|_2^2 + \mathbb{E}_{P_i}\|m_i(X)-g(X_S)\|_2^2$. Consequently, within the restricted class \(\mathcal G_{i,S}\), the subset-dependent part of the squared risk is exactly the error of approximating \(m_i(X)\) using \(X_S\).
\end{theorem}

Thus, subsets differ only through how well \(X_S\) preserves \(m_i(X)\). Because \(m_i\) is unknown, we fit a full-feature teacher \(T_i:\mathcal X\to\mathbb R^q\), treat it as fixed after training, and assume \(T_i(X)\) is square-integrable. Define the teacher-compression risk $C_i^T(S)=\inf_{g\in\mathcal G_{i,S}}\mathbb{E}_{P_i}\ell_T\{T_i(X),g(X_S)\}$, where \(\ell_T\) is the teacher-matching loss.

For squared teacher loss, define
\begin{align*}
D_i^{\mathcal G}(S)
&=
\inf_{g\in\mathcal G_{i,S}}
\mathbb{E}_{P_i}\left\|m_i(X)-g(X_S)\right\|_2^2,\\
\epsilon_i
&=
\mathbb{E}_{P_i}\left\|T_i(X)-m_i(X)\right\|_2^2.
\end{align*}

\begin{theorem}[Teacher-compression bridge]
\label{thm:teacher_bridge}
Under the conditions above, if \(\ell_T(a,b)=\|a-b\|_2^2\), then, conditional on the fitted teacher and with expectations over a fresh draw from \(P_i\), for every population \(i\) and subset \(S\), $\left|\sqrt{C_i^T(S)}-\sqrt{D_i^{\mathcal G}(S)}\right|\leq\sqrt{\epsilon_i}$. Thus, teacher-compression risk approximates full-signal risk on root-risk scale when teacher error is small.
\end{theorem}

Together, Theorems~\ref{thm:projection} and~\ref{thm:teacher_bridge} justify teacher-compression risk as a squared-loss surrogate when the teacher error is small.The bridge controls root risks. In particular, for a fixed population \(i\), if $\sqrt{D_i^{\mathcal G}(S_2)}-\sqrt{D_i^{\mathcal G}(S_1)} > 2\sqrt{\epsilon_i}$, then \(C_i^T(S_1)<C_i^T(S_2)\). 

\subsection{Tuning the Welfare Objective}
\label{sec:tuning_welfare}

A shared feature set may provide substantial benefit to some populations but little benefit to others. We therefore need to specify how the selection objective should trade off aggregate performance against stronger protection of the populations that benefit least. Let \(u_i(S)>0\) denote the utility population \(i\) receives from subset \(S\), and let \(u(S)=(u_1(S),\ldots,u_r(S))\). For population weights \(\rho_i>0\) satisfying \(\sum_{i=1}^r\rho_i=1\), we aggregate these utilities using the weighted power mean
\[
A_{\alpha,\rho}(u)=
\begin{cases}
\left(\sum_{i=1}^r\rho_i u_i^\alpha\right)^{1/\alpha},
& \alpha\neq0,\\
\prod_{i=1}^r u_i^{\rho_i},
& \alpha=0.
\end{cases}
\]
The weights \(\rho_i\) specify the relative importance of the populations, while \(\alpha\) controls how strongly the objective prioritizes populations with lower utility. In particular, \(\alpha=1\) gives the weighted average, \(\alpha=0\) the geometric mean, and \(\alpha\to-\infty\) the minimum. Thus, \(\alpha\) indexes a continuous family of objectives; as it decreases from \(1\), the objective shifts from the weighted average toward minimum-population utility on the floored utility scale. Unless stated otherwise, our experiments use uniform population weights, \(\rho_i=1/r\). Our feature-selection objective is
\begin{align}
S^\star
\in
\arg\max_{\substack{S\subseteq[p]\\|S|=k}}
A_{\alpha,\rho}\left(u(S)\right).
\label{eq:ideal_welfare}
\end{align}

To instantiate \(u_i(S)\), utilities must be comparable across populations with different risk scales and baseline predictability, and strictly positive for the power mean. We therefore measure teacher- or label-based improvement relative to each population's own constant-prediction baseline. Assume that the losses are nonnegative and that all risks below are finite. For direct label-based selection, define \(C_i^Y(S) = \inf_{g\in\mathcal H_{i,S}} \mathbb E_{P_i}\ell_Y\{Y,g(X_S)\}\),
where \(\mathcal H_{i,S}\) is the population-specific label-predictor class. In our experiments, \(\ell_T\) is squared loss; \(\ell_Y\) is squared loss for regression and binary log loss during classification selection, while final classification evaluation uses Brier score. For \(R\in\{T,Y\}\), define
\[
\begin{aligned}
s_i^R(S)
&=
\frac{C_i^R(\emptyset)-C_i^R(S)}
{\max\{C_i^R(\emptyset),\varepsilon_0\}},
\\
u_i^R(S)
&=
\max\{s_i^R(S)+\delta_0,\delta_0\}.
\end{aligned}
\]
Here \(\varepsilon_0,\delta_0>0\), and \(C_i^R(\emptyset)\) is the population risk of the best no-feature constant predictor in the corresponding class. The displayed quantities define the population objective. In computation, \(\widehat C_i^R(S)\) is the unpenalized validation loss of a student fitted on the training partition; the no-feature predictor is likewise fitted on training data and evaluated on validation data. Empirical utilities replace \(C_i^R\) by \(\widehat C_i^R\) throughout. Thus, \(s_i^R(S)\) is normalized improvement over that baseline, while the floor makes \(u_i^R(S)\) strictly positive and treats all nonpositive gains equally. Selection uses \(u_i=u_i^T\) for the teacher objective, \(u_i=u_i^Y\) for the label objective, and \(u_i=\beta u_i^T+(1-\beta)u_i^Y\) for the mixed objective, where \(\beta\in[0,1]\). For the mixed objective, the teacher- and label-target students are fitted separately when computing the two utilities.

Theorem~\ref{thm:alpha_weights} formalizes how decreasing \(\alpha\) increases the relative marginal importance of lower-utility populations.

\begin{theorem}[Lower-utility weighting]
\label{thm:alpha_weights}
For \(u\in(0,\infty)^r\) and finite \(\alpha\in\mathbb R\), the power mean satisfies $\frac{\partial A_{\alpha,\rho}}{\partial u_i}(u)=\rho_i u_i^{\alpha-1}A_{\alpha,\rho}(u)^{1-\alpha}$, where the expression at \(\alpha=0\) is defined by continuity. Hence, as \(d\to0\) with \(u+d\in(0,\infty)^r\), $A_{\alpha,\rho}(u+d)-A_{\alpha,\rho}(u)=\sum_{i=1}^r\rho_i u_i^{\alpha-1}A_{\alpha,\rho}(u)^{1-\alpha}d_i+o(\|d\|_2)$. For any populations \(i\) and \(h\), $\frac{\partial A_{\alpha,\rho}/\partial u_i}{\partial A_{\alpha,\rho}/\partial u_h}=\frac{\rho_i}{\rho_h}\left(\frac{u_i}{u_h}\right)^{\alpha-1}$. Therefore, when \(\rho_i=\rho_h\) and \(\alpha<1\), the population with lower utility receives greater marginal weight.
\end{theorem}

Because a general power mean need not preserve submodularity, even when the population utilities are monotone submodular, our validated greedy-and-swap search is a heuristic for~\eqref{eq:ideal_welfare}; we do not claim a universal constant-factor guarantee for arbitrary \(\alpha\) \citep{das2011submodular,khanna2017scalable}.

\subsection{Teacher Fitting and Candidate Screening}
\label{sec:screening}

To make the subsequent hard-subset search tractable, we construct \(C_{\mathrm{scr}}\) in two screening rounds after fitting population-specific teachers, with \(1\le k\le d\le p_0\le p\). First, an inexpensive marginal-association screen reduces the original \(p\) features to a broad pool \(C_0\) of size \(p_0\). Second, a multitask sparse model considers these features jointly across populations and retains \(d\) features as \(C_{\mathrm{scr}}\), allowing population-specific effects while encouraging shared support.

For each population \(i\), we fit a full-feature teacher \(T_i\) once using all input columns. The teacher supplies the targets for both screening rounds and for teacher-based restricted-model fitting; \method{} does not differentiate through \(T_i\) or its training procedure. In our implementation, \(T_i(x)\) is a scalar prediction---a regression prediction or positive-class probability---so the teacher may be any supported black-box predictor. Teachers are fitted only on the training partition. Their in-sample training predictions (not cross-fitted predictions) are used for screening and feature-opening scores. Candidate students are fitted on training data, move acceptance and final subset selection use unpenalized validation welfare, and the test partition is reserved for final evaluation.

For the first screening round, let
\(\bar x_{ij}=n_i^{-1}\sum_{t=1}^{n_i}x_{itj}\) and
\(\bar z_i=n_i^{-1}\sum_{t=1}^{n_i}T_i(x_{it})\). We score each feature by its marginal association with the teacher across populations:
\begin{align*}
q_j = \left[ \frac{1}{r}\sum_{i=1}^r \left\{ \frac{1}{n_i}\sum_{t=1}^{n_i} (x_{itj}-\bar x_{ij})
\bigl(T_i(x_{it})-\bar z_i\bigr) \right\}^{2} \right]^{1/2}.
\end{align*}
Thus, \(q_j\) is the root-mean-square centered feature--teacher covariance across populations~\citep{fan2008sure}. We retain the \(p_0\) highest-scoring features to form \(C_0\).

For the second round, let \(X_{i,C_0}^{c}\) be population \(i\)'s column-centered design matrix restricted to \(C_0\), and let \(z_i^{c}\) be its centered teacher-output vector. For \(\lambda_{\mathrm{mt}}>0\), we fit the joint sparse model
\begin{align*}
\widehat B
\in
\arg\min_{B\in\mathbb R^{|C_0|\times r}}
\sum_{i=1}^r
\frac{1}{2rn_i}
\left\|
z_i^{c}-X_{i,C_0}^{c}B_{\cdot i}
\right\|_2^2
\\
+
\lambda_{\mathrm{mt}}
\sum_{j\in C_0}
\left\|B_{j\cdot}\right\|_2 .
\end{align*}
The group penalty allows population-specific coefficients while encouraging common support. The \(d\) features in \(C_0\) with the largest row norms \(\|\widehat B_{j\cdot}\|_2\) form \(C_{\mathrm{scr}}\), with \(q_j\) used to break row-norm ties. Because the first round uses marginal feature--teacher covariance, it can discard a jointly useful feature with zero marginal covariance, as may occur for pure interactions. We therefore choose \(p_0\) and \(d\) generously: screening is intended only for high-recall dimension reduction, and subsequent search cannot recover a feature omitted from \(C_{\mathrm{scr}}\).

\subsection{Welfare-Guided Subset Search}

Let \(d = |C_{\mathrm{scr}}|\), and let
\(S_{\mathrm{screen}}\) contain the \(k\) highest-ranked features in
\(C_{\mathrm{scr}}\). PopFS refines this initialization using a
validated one-swap local search. At each iteration, for every
\(h \in S\), we form the base set $B = S \setminus \{h\}$
and consider replacements \(j \in C_{\mathrm{scr}} \setminus S\).
Fully refitting all \(k(d-k)\) possible swaps in every population
would be expensive, so we first rank them using an inexpensive
population-specific surrogate and fully evaluate only a shortlist.

\paragraph{Surrogate ranking.}
For target \(R \in \{T,Y\}\), let \(r_{i,B}^{R} \in \mathbb{R}^{n_i}\)
denote the residual vector of a population-\(i\) ridge ranking model
fitted using \(B\). The target is the teacher prediction when \(R=T\)
and the observed label when \(R=Y\). For a centered candidate column
\(x_{ij} \in \mathbb{R}^{n_i}\) and \(\lambda_{\mathrm{ridge}}>0\), define the auxiliary one-dimensional
ridge objective
\begin{align*}
\psi_{i,j \mid B}^{R}(\gamma)
=
\frac{1}{2n_i}
\left\|
r_{i,B}^{R} - x_{ij}\gamma
\right\|_2^2
+
\frac{\lambda_{\mathrm{ridge}}}{2}\gamma^2 .
\end{align*}
Its predicted reduction from adding feature \(j\) is
\begin{align}
\widetilde{\Delta}_{i}^{R}(j \mid B)
&=
\psi_{i,j \mid B}^{R}(0)
-
\min_{\gamma \in \mathbb{R}}
\psi_{i,j \mid B}^{R}(\gamma)
\nonumber\\
&=
\frac{1}{2}
\frac{
\left(
n_i^{-1}x_{ij}^{\top}r_{i,B}^{R}
\right)^2
}{
n_i^{-1}x_{ij}^{\top}x_{ij}
+
\lambda_{\mathrm{ridge}}
}.
\label{eq:ridge-ranking-gain}
\end{align}
Thus, a candidate receives a large score when it is strongly aligned
with the residual signal that remains after fitting the base set.

\begin{theorem}[Quadratic feature-opening gain]
\label{thm:gn_gain}
Consider
\[
Q(\gamma)
=
Q_0
+
g^{\top}\gamma
+
\frac{1}{2}\gamma^{\top}H\gamma,
\qquad H \succ 0,
\]
where \(H\) is symmetric positive definite. Then the unique minimizer is
\(\gamma^{\star}=-H^{-1}g\), and \(Q(0)-Q(\gamma^{\star}) = \frac{1}{2}g^{\top}H^{-1}g.\)
\end{theorem}

Equation~\eqref{eq:ridge-ranking-gain} is the scalar ridge
specialization of this damped quadratic feature-opening calculation.
The Appendix gives the corresponding residual--Jacobian
Gauss--Newton form for a general differentiable feature opening. The
score is exact only for the local quadratic surrogate, not for the
fully refitted expanded model, and is therefore used only to rank
candidate swaps. This fixed ridge ranking surrogate may also be used
when the shortlisted subsets are evaluated with nonlinear students.
For binary labels, ranking uses the squared-error ridge surrogate, while fully
refitted candidates are evaluated during selection using probabilistic
predictions and binary log loss; final test reporting below uses Brier
score.

For \(R\in\{T,Y\}\), define the normalized surrogate increment
\[
\widetilde d_i^R(j\mid B)
=
\frac{\widetilde\Delta_i^R(j\mid B)}
{\max\{\widehat C_i^R(\emptyset),\varepsilon_0\}}.
\]
The teacher and label objectives use \(\widetilde d_i^T\) and
\(\widetilde d_i^Y\), respectively, and the mixed objective uses
\(\beta\widetilde d_i^T+(1-\beta)\widetilde d_i^Y\). Let
\(\widetilde{\boldsymbol d}(j\mid B)\) denote the resulting vector, and
let \(\widehat{\boldsymbol u}(B)\) be the positive validation-utility
vector obtained by fitting the base-set students on training data and
evaluating them on validation data. The nonnegative surrogate increment
is only a ranking proxy. In particular, when a base utility is at its
floor, the increment is optimistic because it does not offset the
negative raw gain hidden by that floor. We rank each proposed subset by
its surrogate welfare score
\begin{align*}
\widehat{A}(B \cup \{j\})
=
A_{\alpha,\rho}
\left(
\widehat{\boldsymbol u}(B)
+
\widetilde{\boldsymbol d}(j \mid B)
\right).
\end{align*}
Equivalently, for a fixed base set \(B\), candidates may be ranked by
their surrogate increase over
\(A_{\alpha,\rho}(\widehat{\boldsymbol u}(B))\).
Consequently, the welfare parameter \(\alpha\) and population weights
\(\rho\) determine which population-specific improvements are
prioritized, rather than ranking features only by aggregate loss
reduction.

\paragraph{Validated local search.}
In shortlist mode, for each removed feature \(h\), we retain up to
\(L\) of its highest-ranked replacements and pool the resulting swap
proposals. We then fully refit at most
\(\max\{L,L\lfloor |S|/2\rfloor\}\) of the highest-ranked pooled
proposals. In exhaustive mode, we skip shortlisting and fully refit all
\(k(d-k)\) possible swaps. A full evaluation refits the
population-specific restricted students on the training data and
computes their unpenalized welfare on the validation data. We accept
the proposal with the highest validation welfare only if it exceeds
the current welfare by more than \(\delta_{\mathrm{swap}}\). The
procedure then repeats from the updated subset and terminates when no
validated proposal satisfies this condition or after
\(M_{\mathrm{swap}}\) accepted swaps.

Finally, the refined subset is returned only if its validation
welfare exceeds that of \(S_{\mathrm{screen}}\) plus
\(\delta_{\mathrm{safe}}\); otherwise, PopFS returns
\(S_{\mathrm{screen}}\). Thus, the surrogate controls which swaps are
worth the expense of a full refit, while validation welfare determines
every accepted move and the final deployed subset.

\section{Results}
\label{sec:results}

\paragraph{Experimental setup.}
We evaluate \method{} on eight population definitions across six tasks from five heterogeneous tabular and public-health datasets (Table~\ref{tab:datasets}): ACS income, NHANES child and adult blood lead, HELOC credit risk, UCI Adult income\citep{ding2021retiring}, and COVID-19 symptom-search nowcasting\citep{reinhart2021open}. Tasks span 10--2,110 candidate features and 9,870--883,984 observations; geography, age, sex, race+sex, or credit-file age define 73 populations. Each selector chooses one shared feature set, then a separate downstream model is trained per population. The formulation permits different population-specific model classes; within each reported experiment, we use one common class and hyperparameter setting fitted separately in each population. COVID-19 uses 36,772 state-day observations from 43 states (April 1, 2020--November 13, 2022) to nowcast same-day (\(h=0\)) log-transformed hospital admissions from 422 symptom-search signals at lags \(\{0,1,2,3,7\}\) (2,110 features), with within-state 60/20/20 train/validation/test chronological splits.

We compare \method{} with its screening-only ablation (\method{}-Screen), pooled Lasso and XGBoost selection, population-reweighted DRO variants of both \citep{duchi2021learning,sagawa2019distributionally,hashimoto2018fairness} implemented following \citep{swaroop2026distributionally}, and DRFS \citep{swaroop2026distributionally}. DRFS is reported when selection finishes within 24 hours. All methods use the same splits and population-specific downstream model class, isolating feature-set quality.

Let \(G_i(S)=\ell_i(\emptyset)-\ell_i(S)\) denote held-out label gain in population \(i\), where \(\ell_i(\emptyset)\) and \(\ell_i(S)\) are the losses of the population-specific empty predictor and downstream model trained on \(S\). Higher gain indicates greater predictive value beyond baseline. We report its population mean and minimum, measuring average and least-served-population benefit. Regression uses mean squared error, classification uses Brier score, and results are averaged over three random seeds.

\paragraph{\method{} improves both average and worst-population performance.}
 Table~\ref{tab:main_results} reports the main comparison at \(k=6\) and \(\alpha=0\) using gradient-boosting downstream evaluation models. 
\method{} achieves the highest mean in all 16 population-average and worst-population comparisons on both metrics for every split. Relative to the strongest external baseline, its population-average improvement ranges up to \(21.9\%\), with a median of \(5.0\%\) across the eight splits. The largest external-baseline gain occurs on NHANES child lead, where \method{} reaches \(0.284/0.238\) average/worst-population gain, compared with \(0.233/0.187\) for the strongest external baselines. 
 Additional feature budgets, metrics, and model ablations are reported in the Appendix.


\begin{table*}[hbt!]
  \centering
  \small
  \setlength{\tabcolsep}{1.75pt}
  \renewcommand{\arraystretch}{1.00}
  \begin{tabular*}{\textwidth}{@{\extracolsep{\fill}}l*{7}{c}@{}}
    \toprule
    & \multicolumn{3}{c}{\small Population-balanced}
      & \multicolumn{2}{c}{\small Standard}
      & \multicolumn{2}{c}{\small DRO} \\
    \cmidrule(lr){2-4}\cmidrule(lr){5-6}\cmidrule(lr){7-8}
    Dataset & PopFS & PopFS-Screen & DRFS & Lasso & XGBoost & Lasso & XGBoost \\
    \midrule
    \shortstack[l]{UCI Adult\\(sex)} & \awcell{\beststat{0.067}{0.002}}{\beststat{0.043}{0.004}} & \awcell{\stat{0.061}{0.002}}{\stat{0.041}{0.003}} & \awcell{\stat{0.049}{0.013}}{\stat{0.029}{0.013}} & \awcell{\stat{0.062}{0.003}}{\stat{0.041}{0.004}} & \awcell{\stat{0.063}{0.003}}{\stat{0.041}{0.003}} & \awcell{\stat{0.063}{0.003}}{\stat{0.041}{0.004}} & \awcell{\stat{0.064}{0.001}}{\stat{0.042}{0.003}} \\
    \addlinespace[1.0pt]
    \shortstack[l]{UCI Adult\\(race+sex)} & \awcell{\beststat{0.056}{0.001}}{\beststat{0.021}{0.003}} & \awcell{\stat{0.041}{0.002}}{\stat{0.014}{0.004}} & \awcell{\stat{0.050}{0.005}}{\stat{0.019}{0.001}} & \awcell{\stat{0.053}{0.002}}{\stat{0.020}{0.002}} & \awcell{\stat{0.054}{0.002}}{\stat{0.020}{0.002}} & \awcell{\stat{0.054}{0.002}}{\stat{0.020}{0.002}} & \awcell{\stat{0.054}{0.002}}{\beststat{0.021}{0.002}} \\
    \addlinespace[1.0pt]
    \shortstack[l]{UCI Adult\\(age)} & \awcell{\beststat{0.070}{0.002}}{\beststat{0.021}{0.003}} & \awcell{\stat{0.059}{0.009}}{\stat{0.018}{0.003}} & \awcell{\stat{0.055}{0.005}}{\stat{0.018}{0.001}} & \awcell{\stat{0.064}{0.002}}{\stat{0.019}{0.003}} & \awcell{\stat{0.063}{0.002}}{\stat{0.020}{0.003}} & \awcell{\stat{0.064}{0.002}}{\stat{0.019}{0.002}} & \awcell{\stat{0.064}{0.002}}{\stat{0.019}{0.003}} \\
    \addlinespace[2.2pt]
    \shortstack[l]{HELOC} & \awcell{\beststat{0.041}{0.006}}{\beststat{0.027}{0.006}} & \awcell{\stat{0.035}{0.005}}{\stat{0.019}{0.009}} & \awcell{\stat{0.031}{0.004}}{\stat{0.016}{0.005}} & \awcell{\beststat{0.041}{0.005}}{\stat{0.023}{0.011}} & \awcell{\stat{0.040}{0.004}}{\stat{0.023}{0.008}} & \awcell{\beststat{0.041}{0.005}}{\stat{0.023}{0.011}} & \awcell{\beststat{0.041}{0.003}}{\stat{0.021}{0.007}} \\
    \addlinespace[1.0pt]
    ACS Income & \awcell{\beststat{0.689}{0.059}}{\beststat{0.530}{0.055}} & \awcell{\stat{0.659}{0.062}}{\stat{0.510}{0.051}} & \awcell{\stat{0.092}{0.107}}{\stat{0.049}{0.079}} & \awcell{\stat{0.675}{0.050}}{\stat{0.529}{0.044}} & \awcell{\stat{0.681}{0.055}}{\stat{0.525}{0.063}} & \awcell{\stat{0.675}{0.050}}{\stat{0.529}{0.044}} & \awcell{\stat{0.680}{0.058}}{\stat{0.517}{0.047}} \\
    \addlinespace[1.0pt]
    \shortstack[l]{NHANES adult lead} & \awcell{\beststat{0.236}{0.024}}{\beststat{0.183}{0.030}} & \awcell{\stat{0.229}{0.022}}{\stat{0.175}{0.024}} & \awcell{\stat{0.119}{0.049}}{\stat{0.075}{0.036}} & \awcell{\stat{0.223}{0.024}}{\stat{0.172}{0.025}} & \awcell{\stat{0.222}{0.026}}{\stat{0.169}{0.015}} & \awcell{\stat{0.224}{0.025}}{\stat{0.173}{0.024}} & \awcell{\stat{0.222}{0.026}}{\stat{0.168}{0.025}} \\
    \addlinespace[1.0pt]
    \shortstack[l]{NHANES child lead} & \awcell{\beststat{0.284}{0.010}}{\beststat{0.238}{0.030}} & \awcell{\stat{0.249}{0.014}}{\stat{0.211}{0.029}} & \awcell{\stat{0.189}{0.011}}{\stat{0.144}{0.007}} & \awcell{\stat{0.231}{0.016}}{\stat{0.181}{0.017}} & \awcell{\stat{0.224}{0.021}}{\stat{0.187}{0.020}} & \awcell{\stat{0.233}{0.012}}{\stat{0.183}{0.015}} & \awcell{\stat{0.224}{0.014}}{\stat{0.184}{0.018}} \\
    \addlinespace[1.0pt]
    \shortstack[l]{COVID symptoms} & \awcell{\beststat{0.731}{0.013}}{\beststat{0.333}{0.027}} & \awcell{\stat{0.573}{0.008}}{\stat{0.002}{0.005}} & \missingcell & \awcell{\stat{0.667}{0.016}}{\stat{0.257}{0.081}} & \awcell{\stat{0.617}{0.021}}{\stat{0.000}{0.000}} & \awcell{\stat{0.581}{0.006}}{\stat{0.000}{0.000}} & \awcell{\stat{0.653}{0.051}}{\stat{0.145}{0.205}} \\
    \bottomrule
  \end{tabular*}
  \vspace{2pt}
    \caption{
    Population-average (top) and worst-population (bottom) held-out label gain at \(k=6\) and \(\alpha=0\). Entries are mean \(\pm\) one standard deviation over three seeds; higher is better. Every selector is evaluated using the same population-specific downstream model classes and data splits. Bold marks the highest mean for each metric. A dash indicates that feature selection did not finish within 24 hours.
    }
\label{tab:main_results}
  \label{tab:main}

\end{table*}

\paragraph{The validated hard-subset search improves over screening alone.}
The comparison with \method{}-Screen isolates the contribution of the ridge-surrogate shortlist and validation-refit stage. The full method improves over screening alone on both metrics for all eight population splits. The difference is especially large on COVID-19: the screened subset reaches \(0.573\) average gain but only \(0.002\) worst-state gain, whereas the validated search reaches \(0.731\) and \(0.333\), respectively. This result supports the intended division of labor in \method{}: multitask sparsity provides a scalable, high-recall candidate pool, while fully refitted validation welfare determines the deployed hard subset.

\begin{figure}[hbt!]
    \centering
    \includegraphics[width=1.0\linewidth]{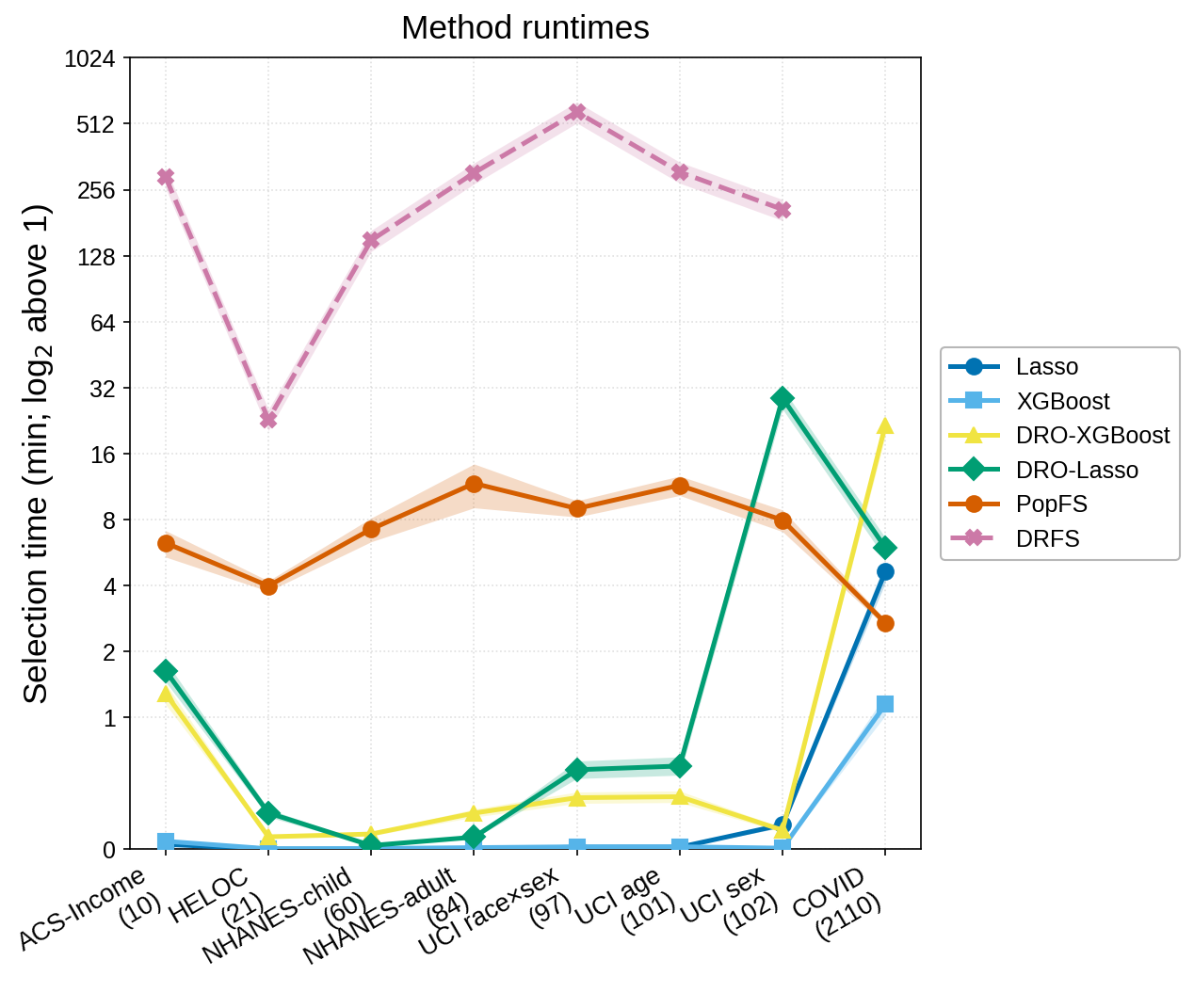}
    \caption{
    End-to-end feature-selection runtime at \(k=6\). The original number of candidate features is shown in parentheses. Lines show means and shaded regions show \(\pm1\) standard error over 3 seeds. All analyses were conducted on an AMD CPU with 16 cores and 64 GB RAM.
    }
    \label{fig:runtime}
\end{figure}

\paragraph{Screening makes discrete search computationally practical.}
Figure~\ref{fig:runtime} reports end-to-end selection time at \(k=6\). \method{} completes every task in approximately 3--12 minutes, including the 2,110-feature COVID-19 task, and is largely insensitive to the total feature count. The sparse screen removes most of the ambient features before the more expensive refit stage, so runtime depends primarily on the screened pool and number of validated candidates rather than on the original dimension. DRFS requires roughly 24--600 minutes on the tasks it completes and does not finish COVID-19 within the 24-hour limit. Thus, the screen--rank--refit strategy enables direct evaluation of deployable hard subsets without exhaustive search or high-dimensional kernel optimization.

\begin{figure}[hbt!]
    \centering
    \includegraphics[width=1.0\linewidth]{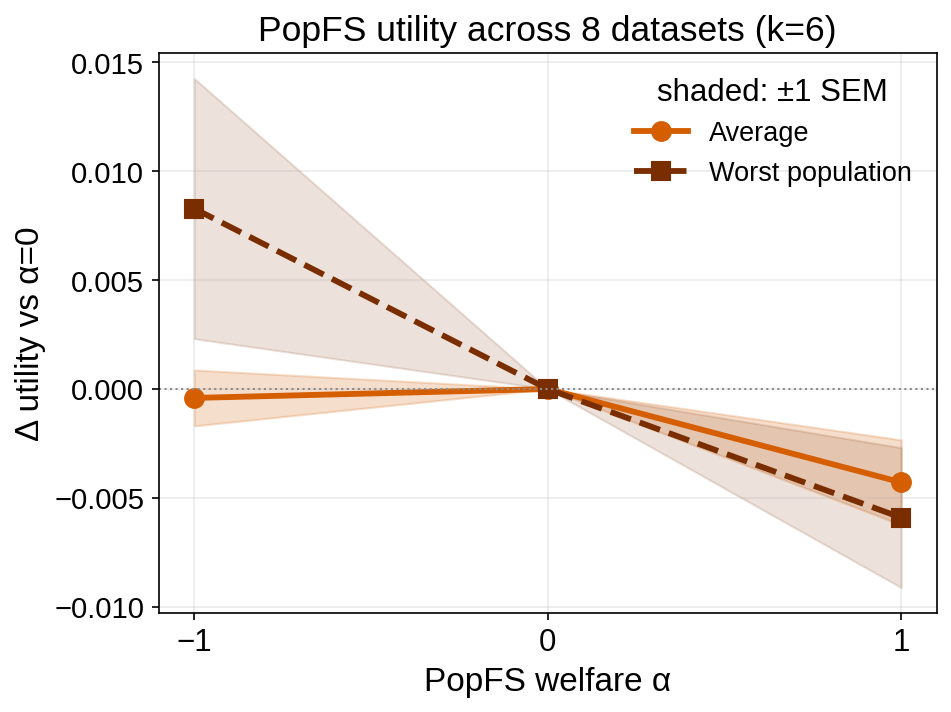}
    \caption{
    Effect of the welfare parameter across eight population splits at \(k=6\). Each point is the mean within-task change in held-out label gain relative to \(\alpha=0\); shaded regions denote \(\pm1\) standard error across population splits.
    }
    \label{fig:alpha_summary}
\end{figure}

\paragraph{The welfare parameter provides meaningful control over who benefits.}
Figure~\ref{fig:alpha_summary} reports the within-task change in test-set label gain relative to \(\alpha=0\), averaged across the eight population splits at \(k=6\). Moving to \(\alpha=-1\) changes population-average gain by less than \(0.001\) on average, while increasing worst-population gain by approximately \(0.01\), or roughly \(11\%\) relative to the corresponding \(\alpha=0\) values. Here, \(\alpha\) is exhibiting its desired behavior of controlling how strongly the acquisition decision should prioritize populations currently receiving lower predictive benefit.

\paragraph{Case study: welfare-aware COVID-19 nowcasting.}

Figure~\ref{fig:covid_case} examines COVID-19 nowcasting at \(k=6\), where one shared symptom-search panel supports separate models across 43 states. In a fixed, separately tuned case-study configuration, decreasing \(\alpha\) from \(1\) to \(-1\) leaves population-average held-out label gain near \(0.64\) but raises worst-state gain from about \(0.16\) to \(0.22\) (about \(40\%\)). Thus, greater emphasis on lower-benefit states improves the lower tail without apparent aggregate cost.

The selected features suggest that timing partly drives this gain. Relative to \(\alpha=0\), the lower-tail setting selects hypoxemia at lag 1 in every additional run, more often selects ageusia at lags 3 and 7 and pneumonia at lag 0, and relies less on other hypoxemia and ageusia lags. Ageusia predicts SARS-CoV-2 infection \citep{menni2020real}, hypoxemic pneumonia characterizes severe COVID-19 \citep{zhang2022human}, and digital traces can precede reported case changes with location-specific timing \citep{kogan2021early}. This may indicate that \method{} replaces nearby, partly redundant lags with a more temporally diverse panel accommodating state-specific lead--lag relationships, although we note \method{} is not designed to detect causal effects.

\begin{strip}
    \centering
    \includegraphics[width=0.9\textwidth]
        {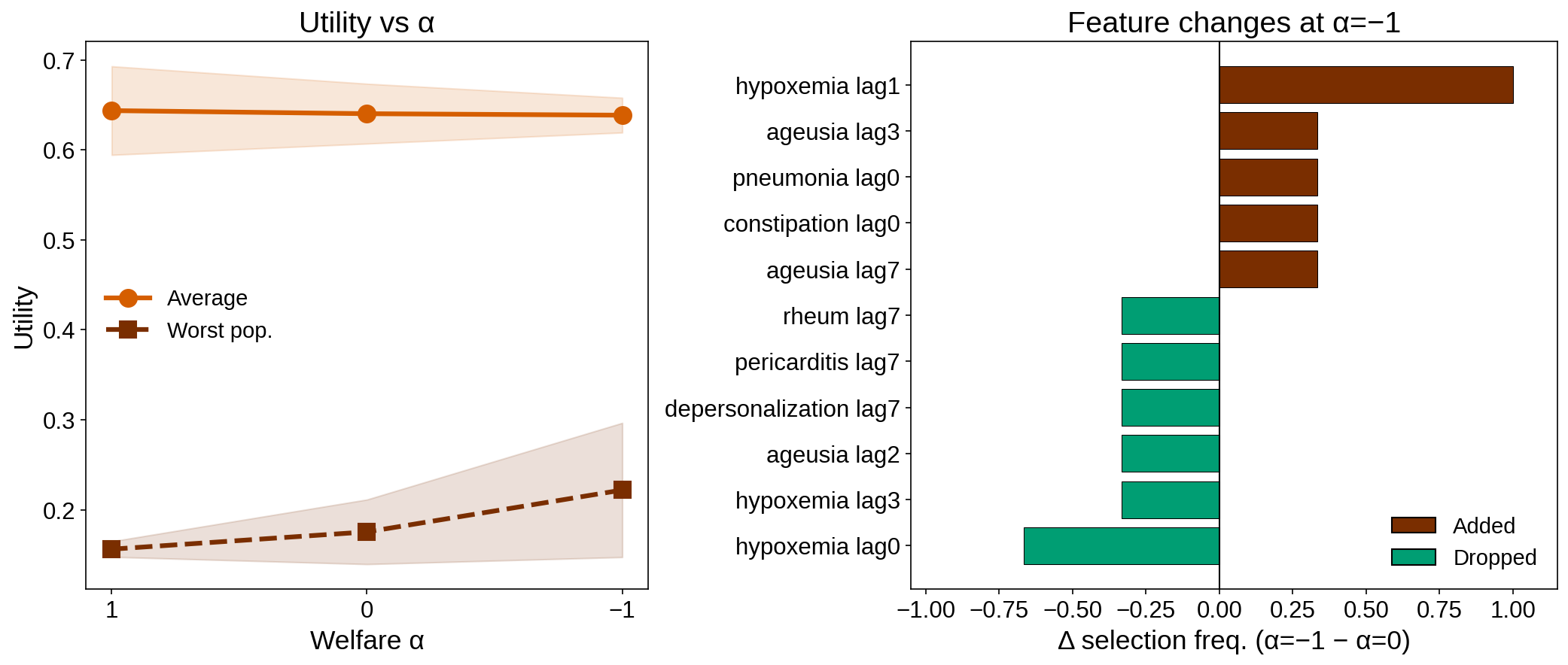}

    \captionof{figure}{
        COVID-19 nowcasting across 43 states at \(k=6\).
        Left: population-average and worst-state held-out label gain
        as the welfare parameter varies; lines show means and shaded
        regions show \(\pm1\) standard error across three seeds.
        Right: change in feature-selection frequency under
        \(\alpha=-1\) relative to \(\alpha=0\).
    }
    \label{fig:covid_case}
\end{strip}

\section*{Discussion}
\label{sec:discussion}

\method{} frames robust feature selection as a shared acquisition decision: one limited feature set must serve heterogeneous populations, even when each population fits its own predictor. \method{} improves both average and worst-population performance, and its welfare parameter balances average and worst-case performance expectedly. We note several limitations and directions for future work, including relying on the quality of the fitted teachers and model classes, and do not imply the heuristic search finds a globally optimal feature set. Future work can extend the framework to unseen or evolving populations and incorporate uncertainty and heterogeneous acquisition costs directly into the selection objective.

\bibliography{aaai2027}


\newpage
\appendix
\onecolumn

\section{Proofs}
\label{app:proofs}

\subsection{Proof of Theorem~\ref{thm:projection}}
\label{app:proof_projection}

We prove the full-signal decomposition. Fix a population \(i\). To simplify
notation, write
\[
    Z=\omega(Y),
    \qquad
    m(X)=m_i(X)=\mathbb{E}\{Z\mid X\},
    \qquad
    h(X_S)=g(X_S).
\]
Because \(Z\) is square-integrable, conditional Jensen's inequality
implies that \(m(X)\) is square-integrable; together with the assumed
square-integrability of \(h(X_S)\), Cauchy--Schwarz then shows that all
expectations below are finite.

We start from the identity
\[
    Z-h(X_S)
    =
    \{Z-m(X)\}+\{m(X)-h(X_S)\}.
\]
Taking squared Euclidean norms and expanding gives
\[
\begin{aligned}
\|Z-h(X_S)\|_2^2
&=
\|Z-m(X)\|_2^2
+
\|m(X)-h(X_S)\|_2^2 \\
&\quad+
2\left\langle
Z-m(X),\,m(X)-h(X_S)
\right\rangle .
\end{aligned}
\]
Now take expectations:
\[
\begin{aligned}
\mathbb{E}\|Z-h(X_S)\|_2^2
&=
\mathbb{E}\|Z-m(X)\|_2^2
+
\mathbb{E}\|m(X)-h(X_S)\|_2^2 \\
&\quad+
2\mathbb{E}
\left[
\left\langle
Z-m(X),\,m(X)-h(X_S)
\right\rangle
\right].
\end{aligned}
\]
It remains to show that the cross term is zero.

Since \(S\subseteq[p]\), \(X_S\) is a function of \(X\). Therefore
\(h(X_S)\) is measurable with respect to \(X\). Also \(m(X)\) is measurable
with respect to \(X\). Hence
\[
    m(X)-h(X_S)
\]
is measurable with respect to \(X\). Using the tower property of conditional
expectation,
\[
\begin{aligned}
&\mathbb{E}
\left[
\left\langle
Z-m(X),\,m(X)-h(X_S)
\right\rangle
\right] \\
&\quad =
\mathbb{E}
\left[
\mathbb{E}
\left\{
\left\langle
Z-m(X),\,m(X)-h(X_S)
\right\rangle
\,\middle|\, X
\right\}
\right].
\end{aligned}
\]
Because \(m(X)-h(X_S)\) is \(X\)-measurable, it can be taken outside the
inner conditional expectation. Thus
\[
\begin{aligned}
&\mathbb{E}
\left\{
\left\langle
Z-m(X),\,m(X)-h(X_S)
\right\rangle
\,\middle|\, X
\right\} \\
&\quad =
\left\langle
\mathbb{E}\{Z-m(X)\mid X\},
m(X)-h(X_S)
\right\rangle .
\end{aligned}
\]
But
\[
    \mathbb{E}\{Z-m(X)\mid X\}
    =
    \mathbb{E}\{Z\mid X\}-m(X)
    =
    m(X)-m(X)
    =
    0.
\]
Therefore the cross term is zero. We obtain
\[
\mathbb{E}\|Z-h(X_S)\|_2^2
=
\mathbb{E}\|Z-m(X)\|_2^2
+
\mathbb{E}\|m(X)-h(X_S)\|_2^2.
\]
Returning to the original notation gives
\[
\mathbb{E}_{P_i}\|\omega(Y)-g(X_S)\|_2^2
=
\mathbb{E}_{P_i}\|\omega(Y)-m_i(X)\|_2^2
+
\mathbb{E}_{P_i}\|m_i(X)-g(X_S)\|_2^2.
\]
Taking the infimum over \(g\in\mathcal G_{i,S}\) on both sides gives
\[
\inf_{g\in\mathcal G_{i,S}}\mathbb E_{P_i}\|\omega(Y)-g(X_S)\|_2^2
=
\mathbb E_{P_i}\|\omega(Y)-m_i(X)\|_2^2+D_i^{\mathcal G}(S).
\]
The first term does not depend on the selected feature subset \(S\), while the
second term is the only term that can depend on \(S\). Hence, under squared loss, the subset-dependent part of the
prediction risk is exactly the error of approximating the full-feature signal
\(m_i(X)\) using only \(X_S\).
\qed

\subsection{Proof of Theorem~\ref{thm:teacher_bridge}}
\label{app:proof_teacher_bridge}

We prove the teacher-compression bridge. Fix a population \(i\) and a subset
\(S\). To simplify notation, write
\[
    m(X)=m_i(X),
    \qquad
    T(X)=T_i(X).
\]
Let
\[
    \mathcal{V}_{i,S}
    =
    \{g(X_S): g\in\mathcal{G}_{i,S}\}
\]
be the set of all predictions that can be produced by the restricted student
class using only \(X_S\).

We use the \(L_2(P_i)\) norm
\[
    \|U\|_{L_2(P_i)}
    =
    \left(\mathbb{E}_{P_i}\|U\|_2^2\right)^{1/2}.
\]

Because the student class is nonempty and, by the stated assumptions, \(T_i(X)\), \(m_i(X)\), and every candidate prediction are square-integrable, the displayed risks are finite. Because the square-root function is increasing and continuous on \([0,\infty)\), taking square roots commutes with the infima below.

With this notation,
\[
    \sqrt{C_i^T(S)}
    =
    \inf_{v\in\mathcal{V}_{i,S}}
    \|T(X)-v\|_{L_2(P_i)}
\]
and
\[
    \sqrt{D_i^{\mathcal G}(S)}
    =
    \inf_{v\in\mathcal{V}_{i,S}}
    \|m(X)-v\|_{L_2(P_i)}.
\]
Also,
\[
    \sqrt{\epsilon_i}
    =
    \|T(X)-m(X)\|_{L_2(P_i)}.
\]

The proof uses only the triangle inequality. For any candidate restricted
prediction \(v\in\mathcal{V}_{i,S}\),
\[
    T(X)-v
    =
    \{T(X)-m(X)\}+\{m(X)-v\}.
\]
Therefore,
\[
    \|T(X)-v\|_{L_2(P_i)}
    \le
    \|T(X)-m(X)\|_{L_2(P_i)}
    +
    \|m(X)-v\|_{L_2(P_i)}.
\]
Substituting the definitions gives
\[
    \|T(X)-v\|_{L_2(P_i)}
    \le
    \sqrt{\epsilon_i}
    +
    \|m(X)-v\|_{L_2(P_i)}.
\]
Now take the infimum over all \(v\in\mathcal{V}_{i,S}\) on both sides:
\[
    \sqrt{C_i^T(S)}
    \le
    \sqrt{\epsilon_i}
    +
    \sqrt{D_i^{\mathcal G}(S)}.
\]
Rearranging,
\[
    \sqrt{C_i^T(S)}-\sqrt{D_i^{\mathcal G}(S)}
    \le
    \sqrt{\epsilon_i}.
\]

We also need the reverse inequality. Again, for any
\(v\in\mathcal{V}_{i,S}\),
\[
    m(X)-v
    =
    \{m(X)-T(X)\}+\{T(X)-v\}.
\]
By the triangle inequality,
\[
    \|m(X)-v\|_{L_2(P_i)}
    \le
    \|m(X)-T(X)\|_{L_2(P_i)}
    +
    \|T(X)-v\|_{L_2(P_i)}.
\]
Since
\[
    \|m(X)-T(X)\|_{L_2(P_i)}
    =
    \sqrt{\epsilon_i},
\]
taking the infimum over \(v\in\mathcal{V}_{i,S}\) gives
\[
    \sqrt{D_i^{\mathcal G}(S)}
    \le
    \sqrt{\epsilon_i}
    +
    \sqrt{C_i^T(S)}.
\]
Rearranging,
\[
    \sqrt{D_i^{\mathcal G}(S)}-\sqrt{C_i^T(S)}
    \le
    \sqrt{\epsilon_i}.
\]

The two inequalities together imply
\[
    \left|
    \sqrt{C_i^T(S)}
    -
    \sqrt{D_i^{\mathcal G}(S)}
    \right|
    \le
    \sqrt{\epsilon_i}.
\]
This proves the result.

The theorem is stated ``conditional on the fitted teacher'' because, after
training, \(T_i\) is treated as a fixed function. The bound says that if the
teacher prediction \(T_i(X)\) is close to the true full-feature signal
\(m_i(X)\) in \(L_2(P_i)\), then teacher-compression risk and full-signal
compression risk are close in root-risk scale.
\qed

\subsection{Proof of Theorem~\ref{thm:alpha_weights}}
\label{app:proof_alpha_weights}

We prove the marginal-weighting formula for the weighted power mean. Let
\[
    u=(u_1,\ldots,u_r),
    \qquad
    u_i>0,
    \qquad
    \rho_i>0,
    \qquad
    \sum_{i=1}^r \rho_i=1.
\]

First consider the case \(\alpha\neq0\). Define
\[
    B_\alpha(u)
    =
    \sum_{i=1}^r \rho_i u_i^\alpha.
\]
Then
\[
    A_{\alpha,\rho}(u)
    =
    B_\alpha(u)^{1/\alpha}.
\]
We compute the derivative with respect to \(u_i\). By the chain rule,
\[
\frac{\partial A_{\alpha,\rho}}{\partial u_i}(u)
=
\frac{1}{\alpha}
B_\alpha(u)^{1/\alpha-1}
\cdot
\frac{\partial B_\alpha}{\partial u_i}(u).
\]
Since
\[
    \frac{\partial B_\alpha}{\partial u_i}(u)
    =
    \rho_i \alpha u_i^{\alpha-1},
\]
we get
\[
\frac{\partial A_{\alpha,\rho}}{\partial u_i}(u)
=
\frac{1}{\alpha}
B_\alpha(u)^{1/\alpha-1}
\cdot
\rho_i\alpha u_i^{\alpha-1}.
\]
Canceling \(\alpha\),
\[
\frac{\partial A_{\alpha,\rho}}{\partial u_i}(u)
=
\rho_i u_i^{\alpha-1}
B_\alpha(u)^{1/\alpha-1}.
\]
Because
\[
    A_{\alpha,\rho}(u)
    =
    B_\alpha(u)^{1/\alpha},
\]
we have
\[
    B_\alpha(u)
    =
    A_{\alpha,\rho}(u)^\alpha.
\]
Therefore
\[
    B_\alpha(u)^{1/\alpha-1}
    =
    \left(A_{\alpha,\rho}(u)^\alpha\right)^{1/\alpha-1}
    =
    A_{\alpha,\rho}(u)^{1-\alpha}.
\]
Substituting back,
\[
\frac{\partial A_{\alpha,\rho}}{\partial u_i}(u)
=
\rho_i u_i^{\alpha-1}
A_{\alpha,\rho}(u)^{1-\alpha}.
\]
This proves the derivative formula for \(\alpha\neq0\).

Now consider the geometric case \(\alpha=0\). The welfare is
\[
    A_{0,\rho}(u)
    =
    \prod_{i=1}^r u_i^{\rho_i}.
\]
Taking logs,
\[
    \log A_{0,\rho}(u)
    =
    \sum_{i=1}^r \rho_i\log u_i.
\]
Differentiating,
\[
    \frac{\partial}{\partial u_i}\log A_{0,\rho}(u)
    =
    \frac{\rho_i}{u_i}.
\]
Using
\[
    \frac{\partial}{\partial u_i}\log A_{0,\rho}(u)
    =
    \frac{1}{A_{0,\rho}(u)}
    \frac{\partial A_{0,\rho}}{\partial u_i}(u),
\]
we obtain
\[
    \frac{\partial A_{0,\rho}}{\partial u_i}(u)
    =
    \rho_i u_i^{-1} A_{0,\rho}(u).
\]
This is exactly the continuous limit of
\[
    \rho_i u_i^{\alpha-1}A_{\alpha,\rho}(u)^{1-\alpha}
\]
as \(\alpha\to0\).

Next, we prove the first-order expansion. Since \(A_{\alpha,\rho}\) is
differentiable for strictly positive \(u_i\), Taylor's theorem gives, as \(d=(d_1,\ldots,d_r)\to0\) with \(u+d\in(0,\infty)^r\),
\[
    A_{\alpha,\rho}(u+d)
    =
    A_{\alpha,\rho}(u)
    +
    \sum_{i=1}^r
    \frac{\partial A_{\alpha,\rho}}{\partial u_i}(u)d_i
    +
    o(\|d\|_2).
\]
Substituting the derivative formula,
\[
A_{\alpha,\rho}(u+d)-A_{\alpha,\rho}(u)
=
\sum_{i=1}^r
\rho_i u_i^{\alpha-1}
A_{\alpha,\rho}(u)^{1-\alpha}
d_i
+
o(\|d\|_2).
\]

Finally, we compare marginal weights between two populations \(i\) and \(h\).
Using the derivative formula,
\[
\frac{
\partial A_{\alpha,\rho}/\partial u_i
}{
\partial A_{\alpha,\rho}/\partial u_h
}
=
\frac{
\rho_i u_i^{\alpha-1} A_{\alpha,\rho}(u)^{1-\alpha}
}{
\rho_h u_h^{\alpha-1} A_{\alpha,\rho}(u)^{1-\alpha}
}.
\]
The common welfare term cancels:
\[
\frac{
\partial A_{\alpha,\rho}/\partial u_i
}{
\partial A_{\alpha,\rho}/\partial u_h
}
=
\frac{\rho_i}{\rho_h}
\left(\frac{u_i}{u_h}\right)^{\alpha-1}.
\]
If \(\rho_i=\rho_h\), this reduces to
\[
\left(\frac{u_i}{u_h}\right)^{\alpha-1}.
\]
Suppose \(u_i<u_h\). Then \(u_i/u_h<1\). If \(\alpha<1\), then
\(\alpha-1<0\), and raising a number smaller than one to a negative power yields a value greater than one:
\[
    \left(\frac{u_i}{u_h}\right)^{\alpha-1}>1.
\]
Thus, when \(\rho_i=\rho_h\) and \(\alpha<1\), the lower-utility population receives greater
marginal weight.
\qed

\subsection{Proof of Theorem~\ref{thm:gn_gain}}
\label{app:proof_gn_gain}
We first prove the generic quadratic-gain formula in
Theorem~\ref{thm:gn_gain}, and then derive the residual--Jacobian
Gauss--Newton form referenced in the method.

\paragraph{Part 1: the local quadratic reduction.}

Fix a current subset \(S\) and a candidate feature \(j\notin S\).
Consider the local quadratic model
\[
    Q(\gamma)
    =
    Q_0
    +
    g^\top\gamma
    +
    \frac12\gamma^\top H\gamma,
    \qquad
    H\succ0.
\]
Here \(g\) and \(\gamma\) have the same dimension, and \(H\) is a
symmetric positive-definite matrix of the corresponding dimension.

To minimize \(Q\), differentiate with respect to \(\gamma\):
\[
    \nabla_\gamma Q(\gamma)
    =
    g+H\gamma.
\]
At the minimizer, the gradient is zero:
\[
    g+H\gamma^\star=0.
\]
Since \(H\succ0\), it is invertible. Therefore
\[
    \gamma^\star=-H^{-1}g.
\]
Now compute the decrease in the quadratic objective:
\[
    Q(0)-Q(\gamma^\star).
\]
First,
\[
    Q(0)=Q_0
\]
Second,
\[
\begin{aligned}
Q(\gamma^\star)
&=
Q_0
+
g^\top(-H^{-1}g)
+
\frac12(-H^{-1}g)^\top H(-H^{-1}g).
\end{aligned}
\]
The middle term is
\[
    g^\top(-H^{-1}g)
    =
    -g^\top H^{-1}g.
\]
The quadratic term is
\[
\begin{aligned}
\frac12(-H^{-1}g)^\top H(-H^{-1}g)
&=
\frac12 g^\top H^{-1} H H^{-1} g \\
&=
\frac12 g^\top H^{-1}g.
\end{aligned}
\]
Thus
\[
\begin{aligned}
Q(\gamma^\star)
&=
Q_0
-
g^\top H^{-1}g
+
\frac12 g^\top H^{-1}g \\
&=
Q_0
-
\frac12 g^\top H^{-1}g.
\end{aligned}
\]
Therefore
\[
    Q(0)-Q(\gamma^\star)
    =
    \frac12 g^\top H^{-1}g.
\]
This proves the quadratic gain formula.

\paragraph{Part 2: residual--Jacobian specialization.}

We now derive the Gauss--Newton quantities for the half-scaled
squared teacher objective used for ranking. Fix a fitted current
predictor \(f_{i,S}(\,\cdot\,;\widehat\theta_{i,S})\). Suppose that the
expanded predictor has an opening parameter
\(\gamma_j\in\mathbb R^{d_j}\), is differentiable in \(\gamma_j\) at
zero, and satisfies
\[
f_{i,S\cup\{j\}}(x_{S\cup\{j\}};
\widehat\theta_{i,S},0)
=
f_{i,S}(x_S;\widehat\theta_{i,S}).
\]
Throughout this calculation, \(\widehat\theta_{i,S}\) is held fixed.

For a training point \(t\), define the current teacher residual
\[
    r_{t,S}
    =
    T_i(x_{it})
    -
    f_{i,S}(x_{it,S};\widehat\theta_{i,S}).
\]
The first-order expansion of the expanded predictor around \(\gamma_j=0\) is
\[
    f_{i,S\cup\{j\}}
    (x_{it,S\cup\{j\}};\widehat\theta_{i,S},\gamma_j)
    \approx
    f_{i,S}(x_{it,S};\widehat\theta_{i,S})
    +
    J_{it,j}\gamma_j,
\]
where
\[
    J_{it,j}
    =
    \left.
    \nabla_{\gamma_j}
    f_{i,S\cup\{j\}}
    (x_{it,S\cup\{j\}};\widehat\theta_{i,S},\gamma_j)
    \right|_{\gamma_j=0}.
\]
For vector-valued outputs, \(J_{it,j}\) is the Jacobian of the output vector
with respect to the opening parameter \(\gamma_j\), and the products
\(J_{it,j}^\top r_{t,S}\) and \(J_{it,j}^\top J_{it,j}\) are interpreted in the
usual vector-Jacobian sense.

Under the linearized expanded model, the teacher residual after opening the
feature is approximately
\[
    r_{t,S}-J_{it,j}\gamma_j.
\]
For \(\lambda_{\mathrm{damp}}>0\), the damped local
half-squared-loss model is
\[
    Q(\gamma_j)
    =
    \frac1{2n_i}
    \sum_{t=1}^{n_i}
    \|r_{t,S}-J_{it,j}\gamma_j\|_2^2
    +
    \frac{\lambda_{\mathrm{damp}}}{2}\|\gamma_j\|_2^2.
\]
We expand the squared norm:
\[
\begin{aligned}
\|r_{t,S}-J_{it,j}\gamma_j\|_2^2
&=
\|r_{t,S}\|_2^2
-
2 r_{t,S}^\top J_{it,j}\gamma_j
+
\gamma_j^\top J_{it,j}^\top J_{it,j}\gamma_j .
\end{aligned}
\]
Substituting into \(Q(\gamma_j)\),
\[
\begin{aligned}
Q(\gamma_j)
&=
\frac1{2n_i}\sum_{t=1}^{n_i}\|r_{t,S}\|_2^2
-
\frac{1}{n_i}\sum_{t=1}^{n_i}
r_{t,S}^\top J_{it,j}\gamma_j \\
&\quad+
\frac1{2n_i}\sum_{t=1}^{n_i}
\gamma_j^\top J_{it,j}^\top J_{it,j}\gamma_j
+
\frac{\lambda_{\mathrm{damp}}}{2}\gamma_j^\top\gamma_j .
\end{aligned}
\]
The constant term is
\[
    Q(0)=\frac1{2n_i}\sum_{t=1}^{n_i}\|r_{t,S}\|_2^2.
\]
The linear term can be written as
\[
    \left(
    -\frac{1}{n_i}
    \sum_{t=1}^{n_i}
    J_{it,j}^\top r_{t,S}
    \right)^\top \gamma_j.
\]
Thus the gradient at zero is
\[
    g_{i,j\mid S}
    =
    -\frac{1}{n_i}
    \sum_{t=1}^{n_i}
    J_{it,j}^{\top}r_{t,S}.
\]

Next, collect the quadratic terms. The curvature matrix multiplying
\(\frac12\gamma_j^\top H\gamma_j\) is
\[
    H_{i,j\mid S}^{\mathrm{GN}}
    =
    \frac{1}{n_i}
    \sum_{t=1}^{n_i}
    J_{it,j}^{\top}J_{it,j}
    +
    \lambda_{\mathrm{damp}}I.
\]
Therefore, up to that additive constant, the local quadratic approximation has the form
\[
    Q(\gamma_j)
    =
    Q(0)
    +
    g_{i,j\mid S}^{\top}\gamma_j
    +
    \frac12
    \gamma_j^\top
    H_{i,j\mid S}^{\mathrm{GN}}
    \gamma_j.
\]
By Part 1, its reduction from opening feature \(j\) is
\[
    \widetilde\Delta_i(j\mid S)
    =
    \frac12
    g_{i,j\mid S}^{\top}
    \left(H_{i,j\mid S}^{\mathrm{GN}}\right)^{-1}
    g_{i,j\mid S}.
\]
This is the corresponding Gauss--Newton feature-opening score.
Because \(\lambda_{\mathrm{damp}}>0\), the curvature is positive
definite. In the scalar linear case \(J_{it,j}=x_{it,j}\), with
\(\lambda_{\mathrm{damp}}=\lambda_{\mathrm{ridge}}\), this expression
reduces exactly to Equation~\eqref{eq:ridge-ranking-gain}.

Finally, note what this calculation does and does not claim. It proves exact
optimality for the local quadratic approximation. It does not claim that this
quantity equals the improvement after fully refitting all parameters of the
expanded student. That is why the main algorithm uses this score only to rank
candidate additions or swaps, and then performs full validation refits before
accepting a move.
\qed

\newpage
\section{Additional experimental results}


\begin{table*}[hbt!]
\centering
\scriptsize
\setlength{\tabcolsep}{3.0pt}
\renewcommand{\arraystretch}{1.05}
\begin{tabular}{lllcrrrrc}
\toprule
Dataset & Prediction target & Population split & Task & $r$ & $p$ & $n$ & Range of $n_i$ & Budgets $k$ \\
\midrule
ACS Income & Income & State & Regression & 10 & 10 & 883,984 & 50,008--195,665 & $\{6,10\}$ \\
NHANES child lead & Blood lead level & Age band & Regression & 3 & 60 & 27,862 & 7,474--10,650 & $\{6,10\}$ \\
NHANES adult lead & Blood lead level & Age band & Regression & 3 & 84 & 47,900 & 11,110--18,559 & $\{6,10\}$ \\
HELOC & Credit risk & Credit-file age tertile & Classification & 3 & 21 & 9,870 & 3,225--3,380 & $\{6,10\}$ \\
UCI Adult (sex) & Income over \$50K & Sex & Classification & 2 & 102 & 30,162 & 9,782--20,380 & $\{6,10\}$ \\
UCI Adult (race+sex) & Income over \$50K & Race+sex & Classification & 6 & 97 & 29,645 & 294--18,038 & $\{6,10\}$ \\
UCI Adult (age) & Income over \$50K & Age tertile & Classification & 3 & 101 & 30,162 & 9,445--10,448 & $\{6,10\}$ \\
COVID-19 symptoms & $\log(1+\mathrm{adm}_{1d})$ ($h=0$) & State & Regression & 43 & 2,110 & 36,772 & 723--950 & $\{6,10\}$ \\
\midrule
\multicolumn{4}{r}{Total population definitions} & 73 & & & & \\
\bottomrule
\end{tabular}
\caption{
Dataset overview. The eight population definitions correspond to six prediction tasks drawn from five datasets. \(n\) and the range of \(n_i\) report post-filter observation counts.
}
\label{tab:datasets}

\end{table*}

\begin{table*}[t]
\centering
\scriptsize
\setlength{\tabcolsep}{4pt}
\renewcommand{\arraystretch}{0.86}
\begin{tabular}{lcccc}
\toprule
& \multicolumn{4}{c}{Teacher / selector model} \\
\cmidrule(lr){2-5}
Dataset & HGB/HGB & HGB/Ridge & Ridge/HGB & Ridge/Ridge \\
\midrule
\multicolumn{5}{l}{\textit{(a) Deployed evaluation student: HGB}} \\
UCI Adult (sex) & \makecell{$\mathbf{0.215\!\pm\!0.007}$\\$\mathbf{0.164\!\pm\!0.005}$} & \makecell{$0.208\!\pm\!0.011$\\$0.161\!\pm\!0.008$} & \makecell{$0.214\!\pm\!0.007$\\$0.163\!\pm\!0.005$} & \makecell{$0.207\!\pm\!0.012$\\$0.160\!\pm\!0.010$} \\
UCI Adult (race+sex) & \makecell{$0.149\!\pm\!0.015$\\$0.011\!\pm\!0.010$} & \makecell{$0.146\!\pm\!0.007$\\$0.006\!\pm\!0.011$} & \makecell{$\mathbf{0.152\!\pm\!0.013}$\\$\mathbf{0.017\!\pm\!0.017}$} & \makecell{$0.142\!\pm\!0.008$\\$0.009\!\pm\!0.009$} \\
UCI Adult (age) & \makecell{$0.207\!\pm\!0.004$\\$0.119\!\pm\!0.001$} & \makecell{$\mathbf{0.209\!\pm\!0.006}$\\$\mathbf{0.121\!\pm\!0.006}$} & \makecell{$0.207\!\pm\!0.004$\\$0.119\!\pm\!0.001$} & \makecell{$\mathbf{0.209\!\pm\!0.006}$\\$\mathbf{0.121\!\pm\!0.006}$} \\
HELOC & \makecell{$0.050\!\pm\!0.022$\\$0.028\!\pm\!0.013$} & \makecell{$0.051\!\pm\!0.009$\\$0.020\!\pm\!0.019$} & \makecell{$0.055\!\pm\!0.015$\\$0.026\!\pm\!0.015$} & \makecell{$\mathbf{0.059\!\pm\!0.005}$\\$\mathbf{0.032\!\pm\!0.018}$} \\
ACS Income & \makecell{$\mathbf{0.251\!\pm\!0.001}$\\$\mathbf{0.226\!\pm\!0.001}$} & \makecell{$0.247\!\pm\!0.001$\\$0.222\!\pm\!0.000$} & \makecell{$\mathbf{0.251\!\pm\!0.001}$\\$\mathbf{0.226\!\pm\!0.001}$} & \makecell{$0.247\!\pm\!0.001$\\$0.222\!\pm\!0.000$} \\
NHANES adult lead & \makecell{$\mathbf{0.118\!\pm\!0.028}$\\$0.076\!\pm\!0.037$} & \makecell{$0.116\!\pm\!0.030$\\$\mathbf{0.080\!\pm\!0.026}$} & \makecell{$0.116\!\pm\!0.030$\\$\mathbf{0.080\!\pm\!0.026}$} & \makecell{$0.116\!\pm\!0.030$\\$\mathbf{0.080\!\pm\!0.026}$} \\
NHANES child lead & \makecell{$\mathbf{0.170\!\pm\!0.032}$\\$0.089\!\pm\!0.022$} & \makecell{$0.160\!\pm\!0.027$\\$0.081\!\pm\!0.024$} & \makecell{$0.159\!\pm\!0.029$\\$\mathbf{0.096\!\pm\!0.036}$} & \makecell{$0.158\!\pm\!0.028$\\$0.073\!\pm\!0.006$} \\
\addlinespace[2pt]
\midrule
\multicolumn{5}{l}{\textit{(b) Deployed evaluation student: Ridge}} \\
UCI Adult (sex) & \makecell{$\mathbf{0.184\!\pm\!0.010}$\\$\mathbf{0.160\!\pm\!0.012}$} & \makecell{$0.180\!\pm\!0.007$\\$0.158\!\pm\!0.009$} & \makecell{$0.181\!\pm\!0.007$\\$0.157\!\pm\!0.009$} & \makecell{$0.180\!\pm\!0.007$\\$0.158\!\pm\!0.009$} \\
UCI Adult (race+sex) & \makecell{$0.151\!\pm\!0.009$\\$0.040\!\pm\!0.025$} & \makecell{$\mathbf{0.156\!\pm\!0.012}$\\$0.042\!\pm\!0.041$} & \makecell{$0.153\!\pm\!0.008$\\$0.047\!\pm\!0.019$} & \makecell{$0.150\!\pm\!0.014$\\$0.041\!\pm\!0.040$} \\
UCI Adult (age) & \makecell{$0.178\!\pm\!0.001$\\$\mathbf{0.112\!\pm\!0.006}$} & \makecell{$\mathbf{0.180\!\pm\!0.001}$\\$\mathbf{0.112\!\pm\!0.003}$} & \makecell{$0.178\!\pm\!0.001$\\$\mathbf{0.112\!\pm\!0.006}$} & \makecell{$\mathbf{0.180\!\pm\!0.001}$\\$\mathbf{0.112\!\pm\!0.003}$} \\
HELOC & \makecell{$0.104\!\pm\!0.008$\\$0.078\!\pm\!0.002$} & \makecell{$0.107\!\pm\!0.011$\\$\mathbf{0.081\!\pm\!0.004}$} & \makecell{$0.104\!\pm\!0.008$\\$0.078\!\pm\!0.002$} & \makecell{$0.107\!\pm\!0.011$\\$0.080\!\pm\!0.003$} \\
ACS Income & \makecell{$\mathbf{0.188\!\pm\!0.002}$\\$\mathbf{0.171\!\pm\!0.001}$} & \makecell{$\mathbf{0.188\!\pm\!0.002}$\\$0.170\!\pm\!0.001$} & \makecell{$\mathbf{0.188\!\pm\!0.002}$\\$\mathbf{0.171\!\pm\!0.001}$} & \makecell{$\mathbf{0.188\!\pm\!0.002}$\\$0.170\!\pm\!0.001$} \\
NHANES adult lead & \makecell{$0.100\!\pm\!0.014$\\$0.077\!\pm\!0.028$} & \makecell{$\mathbf{0.110\!\pm\!0.019}$\\$\mathbf{0.082\!\pm\!0.029}$} & \makecell{$\mathbf{0.110\!\pm\!0.019}$\\$\mathbf{0.082\!\pm\!0.029}$} & \makecell{$\mathbf{0.110\!\pm\!0.019}$\\$\mathbf{0.082\!\pm\!0.029}$} \\
NHANES child lead & \makecell{$0.111\!\pm\!0.005$\\$0.095\!\pm\!0.005$} & \makecell{$0.114\!\pm\!0.004$\\$0.093\!\pm\!0.012$} & \makecell{$0.105\!\pm\!0.016$\\$0.091\!\pm\!0.017$} & \makecell{$\mathbf{0.118\!\pm\!0.005}$\\$0.095\!\pm\!0.010$} \\
\addlinespace[2pt]
\midrule
\multicolumn{5}{l}{\textit{(c) Deployed evaluation student: Random Forest}} \\
UCI Adult (sex) & \makecell{$\mathbf{0.208\!\pm\!0.004}$\\$\mathbf{0.170\!\pm\!0.006}$} & \makecell{$0.203\!\pm\!0.008$\\$0.167\!\pm\!0.007$} & \makecell{$0.206\!\pm\!0.003$\\$0.168\!\pm\!0.003$} & \makecell{$0.202\!\pm\!0.008$\\$0.166\!\pm\!0.007$} \\
UCI Adult (race+sex) & \makecell{$0.152\!\pm\!0.011$\\$0.031\!\pm\!0.027$} & \makecell{$\mathbf{0.159\!\pm\!0.013}$\\$0.034\!\pm\!0.030$} & \makecell{$0.154\!\pm\!0.011$\\$0.046\!\pm\!0.020$} & \makecell{$0.150\!\pm\!0.016$\\$0.034\!\pm\!0.026$} \\
UCI Adult (age) & \makecell{$0.198\!\pm\!0.002$\\$0.120\!\pm\!0.005$} & \makecell{$\mathbf{0.201\!\pm\!0.004}$\\$\mathbf{0.123\!\pm\!0.002}$} & \makecell{$0.198\!\pm\!0.002$\\$0.120\!\pm\!0.005$} & \makecell{$\mathbf{0.201\!\pm\!0.004}$\\$\mathbf{0.123\!\pm\!0.002}$} \\
HELOC & \makecell{$0.103\!\pm\!0.006$\\$0.078\!\pm\!0.006$} & \makecell{$0.105\!\pm\!0.008$\\$0.083\!\pm\!0.013$} & \makecell{$0.102\!\pm\!0.005$\\$0.078\!\pm\!0.007$} & \makecell{$\mathbf{0.106\!\pm\!0.009}$\\$\mathbf{0.084\!\pm\!0.012}$} \\
ACS Income & \makecell{$\mathbf{0.230\!\pm\!0.001}$\\$\mathbf{0.205\!\pm\!0.001}$} & \makecell{$0.227\!\pm\!0.001$\\$0.201\!\pm\!0.002$} & \makecell{$\mathbf{0.230\!\pm\!0.001}$\\$\mathbf{0.205\!\pm\!0.002}$} & \makecell{$0.227\!\pm\!0.001$\\$0.201\!\pm\!0.002$} \\
NHANES adult lead & \makecell{$\mathbf{0.113\!\pm\!0.025}$\\$\mathbf{0.078\!\pm\!0.025}$} & \makecell{$0.106\!\pm\!0.028$\\$0.068\!\pm\!0.020$} & \makecell{$0.106\!\pm\!0.028$\\$0.068\!\pm\!0.020$} & \makecell{$0.106\!\pm\!0.028$\\$0.068\!\pm\!0.020$} \\
NHANES child lead & \makecell{$\mathbf{0.161\!\pm\!0.019}$\\$0.080\!\pm\!0.019$} & \makecell{$0.147\!\pm\!0.014$\\$0.067\!\pm\!0.011$} & \makecell{$0.157\!\pm\!0.021$\\$\mathbf{0.100\!\pm\!0.027}$} & \makecell{$0.150\!\pm\!0.019$\\$0.074\!\pm\!0.014$} \\
\addlinespace[2pt]
\midrule
\multicolumn{5}{l}{\textit{(d) Deployed evaluation student: ExtraTrees}} \\
UCI Adult (sex) & \makecell{$\mathbf{0.182\!\pm\!0.003}$\\$\mathbf{0.156\!\pm\!0.006}$} & \makecell{$0.181\!\pm\!0.004$\\$\mathbf{0.156\!\pm\!0.006}$} & \makecell{$\mathbf{0.182\!\pm\!0.002}$\\$0.155\!\pm\!0.004$} & \makecell{$0.181\!\pm\!0.004$\\$0.155\!\pm\!0.006$} \\
UCI Adult (race+sex) & \makecell{$0.137\!\pm\!0.007$\\$0.031\!\pm\!0.025$} & \makecell{$\mathbf{0.143\!\pm\!0.010}$\\$0.034\!\pm\!0.030$} & \makecell{$0.134\!\pm\!0.008$\\$0.043\!\pm\!0.012$} & \makecell{$0.137\!\pm\!0.015$\\$0.032\!\pm\!0.023$} \\
UCI Adult (age) & \makecell{$\mathbf{0.177\!\pm\!0.002}$\\$\mathbf{0.110\!\pm\!0.008}$} & \makecell{$\mathbf{0.177\!\pm\!0.002}$\\$\mathbf{0.110\!\pm\!0.004}$} & \makecell{$\mathbf{0.177\!\pm\!0.002}$\\$\mathbf{0.110\!\pm\!0.008}$} & \makecell{$\mathbf{0.177\!\pm\!0.002}$\\$\mathbf{0.110\!\pm\!0.004}$} \\
HELOC & \makecell{$0.102\!\pm\!0.003$\\$0.082\!\pm\!0.001$} & \makecell{$0.107\!\pm\!0.007$\\$\mathbf{0.085\!\pm\!0.002}$} & \makecell{$0.102\!\pm\!0.003$\\$0.083\!\pm\!0.001$} & \makecell{$\mathbf{0.108\!\pm\!0.008}$\\$\mathbf{0.085\!\pm\!0.002}$} \\
ACS Income & \makecell{$\mathbf{0.199\!\pm\!0.002}$\\$\mathbf{0.177\!\pm\!0.001}$} & \makecell{$0.196\!\pm\!0.002$\\$0.173\!\pm\!0.002$} & \makecell{$0.198\!\pm\!0.002$\\$0.176\!\pm\!0.002$} & \makecell{$0.196\!\pm\!0.002$\\$0.173\!\pm\!0.002$} \\
NHANES adult lead & \makecell{$\mathbf{0.121\!\pm\!0.022}$\\$\mathbf{0.084\!\pm\!0.034}$} & \makecell{$0.117\!\pm\!0.021$\\$0.081\!\pm\!0.027$} & \makecell{$0.117\!\pm\!0.022$\\$0.080\!\pm\!0.029$} & \makecell{$0.117\!\pm\!0.022$\\$0.080\!\pm\!0.029$} \\
NHANES child lead & \makecell{$0.147\!\pm\!0.017$\\$0.076\!\pm\!0.033$} & \makecell{$0.123\!\pm\!0.017$\\$0.052\!\pm\!0.048$} & \makecell{$\mathbf{0.149\!\pm\!0.021}$\\$\mathbf{0.096\!\pm\!0.032}$} & \makecell{$0.141\!\pm\!0.015$\\$0.072\!\pm\!0.014$} \\
\bottomrule
\end{tabular}
\caption{
\method{} model ablations at \(k=6\) and \(\alpha=0\). Each cell reports population-average (top) and worst-population (bottom) held-out label utility as mean \(\pm\) one standard deviation over three seeds; higher is better. Column labels give the teacher/selector model, and panels fix the deployed evaluation student. \textsc{HGB} denotes histogram gradient boosting. Bold marks the highest mean within each panel and metric. These runs come from the model-ablation store and use a utility scale that is not numerically aligned with the main \(k=6\) gain table.
}
\label{tab:popfs_model_ablations_k6}
\end{table*}

\begin{table*}[t]
\centering
\scriptsize
\setlength{\tabcolsep}{3.1pt}
\renewcommand{\arraystretch}{0.92}
\begin{tabular}{lccccccc}
\toprule
& \multicolumn{3}{c}{Population-balanced} & \multicolumn{2}{c}{Standard} & \multicolumn{2}{c}{DRO} \\
\cmidrule(lr){2-4}\cmidrule(lr){5-6}\cmidrule(lr){7-8}
Dataset & PopFS & PopFS-Screen & DRFS & Lasso & XGBoost & Lasso & XGBoost \\
\midrule
UCI Adult (sex) & \shortstack{$\mathbf{0.0871\!\pm\!0.0011}$\\$\mathbf{0.1181\!\pm\!0.0040}$} & \shortstack{$0.0925\!\pm\!0.0009$\\$0.1275\!\pm\!0.0037$} & \shortstack{$0.1060\!\pm\!0.0129$\\$0.1429\!\pm\!0.0127$} & \shortstack{$0.0921\!\pm\!0.0020$\\$0.1261\!\pm\!0.0055$} & \shortstack{$0.0914\!\pm\!0.0023$\\$0.1253\!\pm\!0.0075$} & \shortstack{$0.0918\!\pm\!0.0023$\\$0.1252\!\pm\!0.0071$} & \shortstack{$0.0902\!\pm\!0.0015$\\$0.1236\!\pm\!0.0066$} \\
UCI Adult (race+sex) & \shortstack{$\mathbf{0.0837\!\pm\!0.0013}$\\$\mathbf{0.1272\!\pm\!0.0066}$} & \shortstack{$0.0983\!\pm\!0.0023$\\$0.1524\!\pm\!0.0035$} & \shortstack{$0.0901\!\pm\!0.0056$\\$0.1394\!\pm\!0.0145$} & \shortstack{$0.0859\!\pm\!0.0022$\\$0.1322\!\pm\!0.0081$} & \shortstack{$0.0849\!\pm\!0.0027$\\$0.1312\!\pm\!0.0035$} & \shortstack{$0.0852\!\pm\!0.0015$\\$0.1312\!\pm\!0.0081$} & \shortstack{$0.0864\!\pm\!0.0015$\\$0.1337\!\pm\!0.0061$} \\
HELOC & \shortstack{$0.1962\!\pm\!0.0023$\\$0.2037\!\pm\!0.0026$} & \shortstack{$0.2035\!\pm\!0.0068$\\$0.2084\!\pm\!0.0055$} & \shortstack{$0.2032\!\pm\!0.0042$\\$0.2077\!\pm\!0.0042$} & \shortstack{$0.1952\!\pm\!0.0014$\\$\mathbf{0.1982\!\pm\!0.0032}$} & \shortstack{$0.1956\!\pm\!0.0032$\\$0.1984\!\pm\!0.0043$} & \shortstack{$0.1952\!\pm\!0.0014$\\$\mathbf{0.1982\!\pm\!0.0032}$} & \shortstack{$\mathbf{0.1951\!\pm\!0.0036}$\\$0.2009\!\pm\!0.0027$} \\
\midrule
ACS Income & \shortstack{$\mathbf{0.3269\!\pm\!0.0412}$\\$\mathbf{0.4199\!\pm\!0.0503}$} & \shortstack{$0.3645\!\pm\!0.0461$\\$0.4662\!\pm\!0.0790$} & \shortstack{$0.9278\!\pm\!0.1359$\\$1.0597\!\pm\!0.1073$} & \shortstack{$0.3392\!\pm\!0.0304$\\$0.4502\!\pm\!0.0633$} & \shortstack{$0.3377\!\pm\!0.0397$\\$0.4256\!\pm\!0.0725$} & \shortstack{$0.3392\!\pm\!0.0304$\\$0.4502\!\pm\!0.0633$} & \shortstack{$0.3395\!\pm\!0.0293$\\$0.4348\!\pm\!0.0671$} \\
NHANES child lead & \shortstack{$\mathbf{0.7967\!\pm\!0.0128}$\\$0.9040\!\pm\!0.0454$} & \shortstack{$0.8114\!\pm\!0.0241$\\$\mathbf{0.9012\!\pm\!0.0344}$} & \shortstack{$0.8370\!\pm\!0.0212$\\$0.9434\!\pm\!0.0735$} & \shortstack{$0.8274\!\pm\!0.0222$\\$0.9232\!\pm\!0.0559$} & \shortstack{$0.8160\!\pm\!0.0156$\\$0.9230\!\pm\!0.0280$} & \shortstack{$0.8289\!\pm\!0.0197$\\$0.9094\!\pm\!0.0322$} & \shortstack{$0.8122\!\pm\!0.0166$\\$0.9268\!\pm\!0.0445$} \\
\bottomrule
\end{tabular}
\caption{
Held-out predictive loss at $k=6$. Each cell reports population-average (top) and worst-population (bottom) loss as mean $\pm$ one standard deviation over three seeds; lower is better. Regression rows report mean squared error, whereas classification rows report Brier score. Bold marks the lowest mean for each metric.
}
\label{tab:k6_predictive_losses}
\end{table*}

\begin{table*}[t]
\centering
\scriptsize
\setlength{\tabcolsep}{3.1pt}
\renewcommand{\arraystretch}{0.92}
\begin{tabular}{lccccccc}
\toprule
& \multicolumn{3}{c}{Population-balanced} & \multicolumn{2}{c}{Standard} & \multicolumn{2}{c}{DRO} \\
\cmidrule(lr){2-4}\cmidrule(lr){5-6}\cmidrule(lr){7-8}
Dataset & PopFS & PopFS-Screen & DRFS & Lasso & XGBoost & Lasso & XGBoost \\
\midrule
UCI Adult (sex) & \shortstack{$\mathbf{0.2220\!\pm\!0.0023}$\\$\mathbf{0.1717\!\pm\!0.0029}$} & \shortstack{$0.2013\!\pm\!0.0055$\\$0.1538\!\pm\!0.0106$} & -- & \shortstack{$0.2140\!\pm\!0.0025$\\$0.1653\!\pm\!0.0032$} & \shortstack{$0.1951\!\pm\!0.0030$\\$0.1596\!\pm\!0.0056$} & \shortstack{$0.2192\!\pm\!0.0025$\\$0.1678\!\pm\!0.0030$} & \shortstack{$0.2080\!\pm\!0.0082$\\$0.1602\!\pm\!0.0099$} \\
UCI Adult (race+sex) & \shortstack{$0.1428\!\pm\!0.0074$\\$0.0038\!\pm\!0.0034$} & \shortstack{$0.0990\!\pm\!0.0081$\\$0.0164\!\pm\!0.0164$} & -- & \shortstack{$0.1428\!\pm\!0.0059$\\$0.0070\!\pm\!0.0070$} & \shortstack{$0.1420\!\pm\!0.0059$\\$\mathbf{0.0210\!\pm\!0.0109}$} & \shortstack{$\mathbf{0.1553\!\pm\!0.0068}$\\$0.0148\!\pm\!0.0080$} & \shortstack{$0.1395\!\pm\!0.0073$\\$0.0148\!\pm\!0.0148$} \\
UCI Adult (age) & \shortstack{$\mathbf{0.2137\!\pm\!0.0033}$\\$\mathbf{0.1217\!\pm\!0.0017}$} & \shortstack{$0.2045\!\pm\!0.0014$\\$0.1143\!\pm\!0.0004$} & -- & \shortstack{$0.2042\!\pm\!0.0090$\\$0.1196\!\pm\!0.0051$} & \shortstack{$0.1697\!\pm\!0.0191$\\$0.1024\!\pm\!0.0109$} & \shortstack{$0.2082\!\pm\!0.0026$\\$0.1201\!\pm\!0.0016$} & \shortstack{$0.2029\!\pm\!0.0027$\\$0.1144\!\pm\!0.0016$} \\
HELOC & \shortstack{$0.0379\!\pm\!0.0020$\\$0.0224\!\pm\!0.0022$} & \shortstack{$0.0380\!\pm\!0.0004$\\$0.0219\!\pm\!0.0048$} & \shortstack{$0.0368\!\pm\!0.0034$\\$0.0194\!\pm\!0.0041$} & \shortstack{$0.0365\!\pm\!0.0003$\\$0.0248\!\pm\!0.0047$} & \shortstack{$0.0388\!\pm\!0.0021$\\$0.0231\!\pm\!0.0019$} & \shortstack{$0.0365\!\pm\!0.0003$\\$0.0248\!\pm\!0.0047$} & \shortstack{$\mathbf{0.0405\!\pm\!0.0024}$\\$\mathbf{0.0260\!\pm\!0.0009}$} \\
ACS Income & \shortstack{$\mathbf{0.2589\!\pm\!0.0004}$\\$\mathbf{0.2344\!\pm\!0.0010}$} & \shortstack{$\mathbf{0.2589\!\pm\!0.0004}$\\$\mathbf{0.2344\!\pm\!0.0010}$} & -- & \shortstack{$\mathbf{0.2589\!\pm\!0.0004}$\\$\mathbf{0.2344\!\pm\!0.0010}$} & \shortstack{$\mathbf{0.2589\!\pm\!0.0004}$\\$\mathbf{0.2344\!\pm\!0.0010}$} & \shortstack{$\mathbf{0.2589\!\pm\!0.0004}$\\$\mathbf{0.2344\!\pm\!0.0010}$} & \shortstack{$\mathbf{0.2589\!\pm\!0.0004}$\\$\mathbf{0.2344\!\pm\!0.0010}$} \\
NHANES adult lead & \shortstack{$\mathbf{0.1430\!\pm\!0.0179}$\\$\mathbf{0.0928\!\pm\!0.0157}$} & \shortstack{$0.1233\!\pm\!0.0157$\\$0.0879\!\pm\!0.0141$} & -- & \shortstack{$0.1246\!\pm\!0.0162$\\$0.0885\!\pm\!0.0165$} & \shortstack{$0.1182\!\pm\!0.0198$\\$0.0805\!\pm\!0.0182$} & \shortstack{$0.1209\!\pm\!0.0164$\\$0.0848\!\pm\!0.0146$} & \shortstack{$0.1103\!\pm\!0.0133$\\$0.0648\!\pm\!0.0148$} \\
NHANES child lead & \shortstack{$\mathbf{0.1688\!\pm\!0.0161}$\\$0.0837\!\pm\!0.0116$} & \shortstack{$0.1358\!\pm\!0.0104$\\$0.0650\!\pm\!0.0074$} & -- & \shortstack{$0.1586\!\pm\!0.0157$\\$0.0845\!\pm\!0.0152$} & \shortstack{$0.1612\!\pm\!0.0128$\\$\mathbf{0.0896\!\pm\!0.0114}$} & \shortstack{$0.1605\!\pm\!0.0191$\\$0.0779\!\pm\!0.0172$} & \shortstack{$0.1315\!\pm\!0.0114$\\$0.0666\!\pm\!0.0068$} \\
\bottomrule
\end{tabular}
\caption{
Held-out label utility at $k=10$ and $\alpha=0$. Each cell reports population-average (top) and worst-population (bottom) utility as mean $\pm$ one standard error over three seeds; higher is better. Bold marks the highest available mean for each metric, and dashes indicate that an aligned result was unavailable or did not finish within the computational budget. The entries use the $k=10$ $\alpha$-sweep utility scale and are directly comparable within this table, but not numerically aligned with the main $k=6$ gain table. For ACS Income, $k=p=10$, so every available method selects all features and the results coincide.
}
\label{tab:k10_utility}
\end{table*}

\end{document}